\pdfoutput=1

\documentclass[11pt]{article}

\usepackage[final]{acl}

\usepackage{times}
\usepackage{latexsym}
\usepackage{booktabs}
\usepackage{tabularx}
\usepackage{hhline}
\usepackage{arydshln}
\usepackage{pgfplots}
\pgfplotsset{compat=1.17}
\usepackage{boxedminipage}
\usepackage{makecell}

\usepackage[T1]{fontenc}

\usepackage[utf8]{inputenc}

\usepackage{microtype}
\usepackage{multirow}

\usepackage{inconsolata}
\usepackage{CJKutf8}

\usepackage{graphicx}
\usepackage{algorithm}
\usepackage{algorithmic}
\usepackage{amsmath} 
\usepackage{amssymb}

\title{ Chinese Morph Resolution in E-commerce Live Streaming Scenarios}

  \author{
  Jiahao Zhu\textsuperscript{1}  ,
  Jipeng Qiang\textsuperscript{1}\thanks{{\; Corresponding authors. }}, 
  Ran Bai\textsuperscript{2},
  Chenyu Liu\textsuperscript{2}, 
  Xiaoye Ouyang\textsuperscript{2}\\
\textsuperscript{1} School of Information Engineering, Yangzhou University, China\\
\textsuperscript{2} China Academy of Electronic and Information Technology, China
\\ mz120231031@stu.yzu.edu.cn, jpqiang@yzu.edu.cn
\\ \{bairan, liuchenyu, ouyangxiaoye \}@cetc.com.cn
}

\begin{document}
\maketitle
\begin{abstract}
E-commerce live streaming in China, particularly on platforms like Douyin, has become a major sales channel, but hosts often use morphs to evade scrutiny and engage in false advertising. This study introduces the Live Auditory Morph Resolution (LiveAMR) task to detect such violations. Unlike previous morph research focused on text-based evasion in social media and underground industries, LiveAMR targets pronunciation-based evasion in health and medical live streams. We constructed the first LiveAMR dataset with 86,790 samples and developed a method to transform the task into a text-to-text generation problem. By leveraging large language models (LLMs) to generate additional training data, we improved performance and demonstrated that morph resolution significantly enhances live streaming regulation.

\end{abstract}

\section{Introduction}

\begin{figure}[!h]
  \includegraphics[width=\columnwidth]{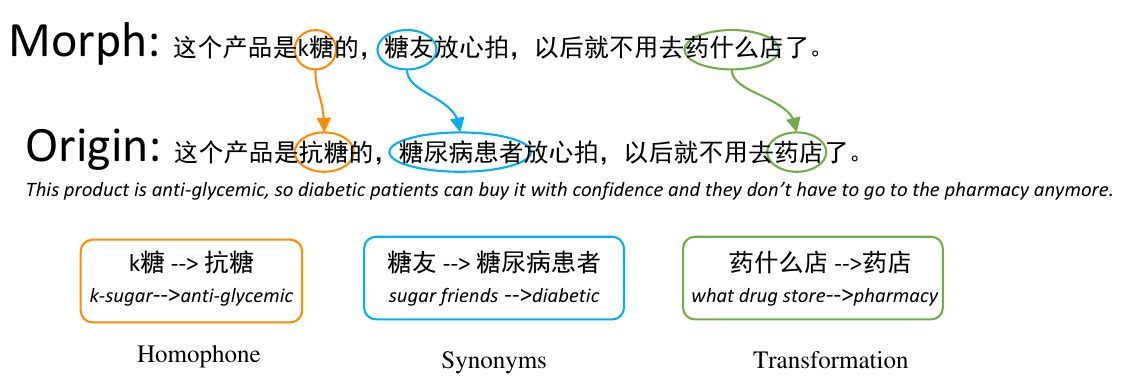}
  \caption{Example of morph used in the live streaming scenarios}
  \label{fig:example entity morph}
\end{figure}

E-commerce live streaming has become an immensely popular and influential sales channel in China. For example, one short video platform Douyin hosted over 9 million live broadcasts each month, selling more than 10 billion items through there sessions \cite{center202250th}. To increase sales and attract customers, hosts engage in practices such as using morphs to evade scrutiny and conducting false advertising. As shown in the Figure \ref{fig:example entity morph}, morphs are used in promotional language that suggests the product has medicinal effects in order to evade scrutiny. Detecting violations during the live commerce process is crucial for protecting consumer rights and promoting industry standardization \cite{xiao2024rise,xu2024}.

To detect violations in live commerce, resolving morphs used in the live content is intuitively important. Previous morph research has primarily focused on social media commentary and underground industries \cite{sha2017resolving,you2018morph,wang2024cmright}. There are two main differences between their research and this paper. 

(1) Different purposes for morphing:  Their focus is on making the written text appear different to evade keyword recognition \cite{you2018morph,wang2024cmright}, whereas the live streaming field focuses on differences in pronunciation to evade voice censorship.  For example, in visual scenarios, characters with a left-right structure are often split into two words, such as \begin{CJK}{UTF8}{gbsn}‘‘胡’’ 
(\textit{h\'{u}})\end{CJK}->\begin{CJK}{UTF8}{gbsn}‘‘古月’’(\textit{g\v{u} y\`{u}e})\end{CJK}. In the live streaming field, a very common situation is inserting some meaningless words , like \begin{CJK}{UTF8}{gbsn}‘’某’’(\textit{m\v{o}u, some})\end{CJK} or \begin{CJK}{UTF8}{gbsn}‘‘什么’’(\textit{sh\'{e}n m\={e}, what})\end{CJK} can help maintain the rhythm of speech without interfering with the listener's understanding of the information, such as \begin{CJK}{UTF8}{gbsn}‘‘手术’’(\textit{sh\v{o}u sh\`{u}, surgery})->‘‘手某术’’(\textit{sh\v{o}u m\v{o}u sh\`{u}, surgery})\end{CJK}". 

(2) Different subjects of interest: Social media commentary focuses on current affairs and politics \cite{you2018morph}, and underground industries focus on illegal gambling and the sex industry \cite{wang2024cmright}, while our study focuses on the health and medical industry.

In this paper, we focus on auditory-based morph resolution task in live screaming scenarios, denoted as LiveAMR task. Voice censorship is first processed using automatic speech recognition (ASR) technology \cite{wang2023complex}, which converts speech into text. By observation, we can find that the LiveAMR task is similar to the grammar correction task \cite{kobayashi2024revisiting}. In this way, we can train a text generation model to convert the input text with morph words into normal text. This study produces two main contributions toward the development and evaluation of LiveAMR methods. Our contributions are listed below:

(1) To the best of our knowledge, there is no existing work on LiveAMR. We constructed a LiveAMR dataset containing 86,790 samples, including 2,688 different morphs. In live streaming scenarios, considering the noise in the live environment and the variations in presenters' expressions, the results of different ASR systems vary greatly. We re-annotated the second test set, selecting different live streaming rooms and different ASR methods which includes 400 positive and 400 negative samples. This approach allows us to comprehensively assess the model's performance and adaptability under different conditions.

(2) We transform LiveAMR task into a type of text-to-text generation task. By training the T5 model using the constructed morph dataset, we achieved F1 scores of 94\% and 82\% on Test Set 1 and Test Set 2, surpassing the performance of other models respectively. Considering the efficiency of manual annotation is relatively low, we propose an innovative solution that leverages large language modeling to generate LiveAMR examples, thereby improving the scale of LiveAMR training set. Experimental results show that incorporating the dataset generated by LLM into the training process also improved the performance of LiveAMR methods. Additionally, we investigated the peformance of morph resolution in detecting violations. We also verify that morph resolution can significantly improve the model's accuracy in the live streaming regulation. The dataset and code is available at github \footnote{https://github.com/loopback00/LiveAMR}.


\section{Related Work}

There has been extensive research on morph resolution across different language backgrounds including English \cite{ji2018creative,li2022name,wang2023mttm,qiang2023parals}, and Chinese \cite{huang2017kiem,huang2019multimodal,qiang2023chinese}, etc. In this paper, we only focus on morph resolution in Chinese. Because Chinese is a pictographic language, methods for identifying morph words in other languages cannot be applied to Chinese. Existing research on Chinese morphs primarily focuses on social media and underground industries.  

Initially, it was considered a filtering problem, with researchers using statistical and rule-based matching methods to identify problematic text \cite{wang2013chinese,choudhury2007investigation,qianglexi2023chinese,yoon2010smart}. Subsequently, Sha et al. \cite{sha2017resolving} proposed incorporating radicals into Chinese characters to enhance their features and improve morphs resolution. You et al.\cite{you2018morph} further extracted actual contextual information and enhanced embedded representations by integrating transformed mentions or target candidates with their relevant context into an AutoEncoder. Recently, addressing the characteristics of morph words in underground industries, Wang et al.\cite{wang2024cmright} introduced a morph parsing algorithm based on machine translation models.

However, existing research on morphs mainly focuses on social media and underground industries, with studies on morph resolution in the emerging context of live streaming still being relatively scarce.

\section{Task Definition}
\label{section:task}

\begin{table*}[!t]
\small
    \centering
    \begin{tabularx}{16cm}{m{2.5cm}m{8cm}m{5.5cm}}
        \toprule
        
        \textbf{Type} &
        \centering
        \textbf{Characteristic} & 
        
        \textbf{Examples}
        \\
        \midrule
        Transformation&
       Insert meaningless characters into words, or change the structure while keeping the sound similar to the original words.&
       \begin{CJK}{UTF8}{gbsn}某医某院\end{CJK}:\begin{CJK}{UTF8}{gbsn}医院\end{CJK} (hospital) \newline
       \textit{m\v{o}u y\={i} m\v{o}u yu\`{a}n:y\={i} yu\`{a}n}
       \newline
       \begin{CJK}{UTF8}{gbsn}祛什么斑\end{CJK}:\begin{CJK}{UTF8}{gbsn}祛斑\end{CJK} (spot removal) \newline
       \textit{q\={u} sh\'{e}n m\={e} b\={a}n:q\={u} b\={a}n}
       \newline
       \begin{CJK}{UTF8}{gbsn}小问小题\end{CJK}:\begin{CJK}{UTF8}{gbsn}问题\end{CJK} (problem)
       \newline
       \textit{xi\v{a}o w\`{e}n xi\v{a}o t\'{i}:w\`{e}n t\'{i}}
       \\
       \hline
        Homophone
        &Use symbols to replace Chinese characters
        &
        \begin{CJK}{UTF8}{gbsn}k糖\end{CJK}:\begin{CJK}{UTF8}{gbsn}抗糖\end{CJK} (anti glycemic) \newline
        \textit{k t\'{a}ng: k\`{a}ng t\'{a}ng}
        \newline
        \begin{CJK}{UTF8}{gbsn}k老\end{CJK}:\begin{CJK}{UTF8}{gbsn}抗老\end{CJK} (anti aging)
        \newline
        \textit{k l\v{a}o: k\`{a}ng l\v{a}o}
        \\
        \hline
        Synonyms&
        Use words that are highly related or synonymous with the target word
        &
          \begin{CJK}{UTF8}{gbsn}白大褂\end{CJK}:\begin{CJK}{UTF8}{gbsn}医生\end{CJK}  
          \newline
          (people in white:doctor)
          \newline
        \begin{CJK}{UTF8}{gbsn}心灵之窗\end{CJK}:\begin{CJK}{UTF8}{gbsn}眼睛\end{CJK} \newline (windows to the soul:eyes)
        \\ 
       \bottomrule
    \end{tabularx}
    \caption{The three types of transformations in LiveAMR. For the two types of morphs, transformation and homophone, we have additionally annotated their pinyin below them.}
    \label{tab:1}
\end{table*}

In the research context of this paper, 'morph' refers to the process where live streamers avoid platform censorship by replacing sensitive or restrictive words during product promotion, while ensuring that the audience can easily understand the original meaning conveyed by the transformation. Here, we formally define the auditory-based morph resolution task in live screaming scenarios as the LiveAMR task. By analyzing thousands of videos, the main types of transformations can be categorized into three major types (transformation, homophones, and synonyms), as shown in Table \ref{tab:1}.

Suppose one example is  "\begin{CJK}{UTF8}{gbsn}咱们一些<小糖人>都是一样可以放心去喝，也不用去找<白褂褂>了。\end{CJK}" (Some diabetes patients can safely drink without needing to consult a doctor.) with two morphs \begin{CJK}{UTF8}{gbsn}‘‘小糖人’’(\textit{sugar doll})->‘‘糖尿病患者’’(\textit{diabetic})\end{CJK} and \begin{CJK}{UTF8}{gbsn}‘‘白褂褂’’(\textit{people with white})->‘‘医生’’(\textit{ doctor})\end{CJK}. The correct output by LiveAMR method should be “\begin{CJK}{UTF8}{gbsn}咱们一些<糖尿病患者>都是一样可以放心去喝，也不用去找<医生>了。\end{CJK}”.

\section{Dataset Construct}

In this section, we describe the whole process of constructing a LiveAMR dataset.

\textbf{Data Collection:}  We crawled videos from four domains in Douyin website \footnote{https://www.douyin.com/}: health supplements, pharmaceuticals, medical devices, and cosmetics. These areas are chosen due to their unique risks and challenges in live streaming. As products aimed at improving health, they have a large market size and diverse categories. However, due to their specific nature, consumers often face significant information asymmetry regarding their efficacy and safety. This asymmetry creates opportunities for false advertising and misleading marketing, particularly in the highly interactive and instant-feedback environment of live streaming \cite{auronen2003asymmetric}.

From the four domains, we carefully selected 25 live streaming channels as data sources. These channels are well-known on the platform and have high sales, ensuring they are representative. We crawled a total of 7,812 live video clips, each limited to 60 seconds. This duration ensures sufficient information capture while reducing data processing complexity to some extent, providing rich material for subsequent data annotation.

\textbf{ASR Process:} We first need to convert the audio information into text format. We tested the transcription performance of mainstream ASR tools in this scenario, with FunASR \cite{gao2023funasr} achieving the best recognition results, followed by Kaldi \cite{ravanelli2019pytorch} and Whisper \cite{radford2023robust}. We employed this FunASR to perform ASR, converting the spoken content in the crawled videos into text for subsequent morph annotation. A total of 86,750 speech statements were transcribed.

This process of converting video to text not only adds a new modality to the research but also makes the form of morphs more flexible and varied. In the video context, morph words themselves are very difficult to distinguish by ASR. Additionally, other factors such as the host's colloquial expressions, fast speaking pace, and background noise can lead to inaccuracies in ASR recognition results, resulting in a more diverse range of extracted morph forms.

\textbf{Label Suggestions via LLMs:}  Recently, LLMs have been widely used for data annotation \cite{zhang-etal-2023-llmaaa}. Despite the challenges posed by the presence of grammatical morphs in the annotation of  morphs, LLMs with their powerful contextual learning capabilities, can still identify some standard morphs and provide the correct original terms. Therefore, we provided the annotation suggestions from the LLMs to human annotators as a reference, assisting them in the annotation process to enhance both efficiency and accuracy. Whether some morphs recommended by the LLMs actually exist in the original document, annotators can more quickly locate the variant words. To specifically illustrate the performance of LLMs in LiveAMR task, we selected three representative LLMs as baselines to comparison.

\textbf{Human Annotation:}  In order to make it easier for annotators to label, we created a website for annotation. We provided corresponding videos and LLM annotation suggestions as auxiliary information, with video support being essential. When we attempted annotation without referencing the videos, annotators reported that many words could not be clearly understood. We recruited three interns with bachelor's degrees with annotation experience and an understanding of morph characteristics as annotators. 

The unique research scenarios required annotators to process multiple modalities of information, enhancing the quality and accuracy of the annotations. Prior to formal annotation, detailed training was provided, including explanations of guidelines and procedures, along with trial annotations to ensure understanding and adherence to the tasks. Each annotator needs to undergo training before starting their annotation work, and they can only begin once they have passed the training. As a result, the annotation process yielded 6,853 positive sentences containing morphs and 90,137 negative sentences without morphs.

\textbf{Data Filtering:} Despite manually annotating morph words, we found that a small number of variant words were still not annotated. Therefore, we further adopted a process of human-machine collaboration for secondary annotation to achieve the goal of constructing a high-quality dataset.

First, we use the corpus manually annotated in the previous step to build a morph resolution model, employing both rule-based method and pre-trained language model based method. Second, we automatically annotate the manually annotated corpus from the previous step using the trained method. Third, we manually verify the correctness of the machine's automatic annotation results, retaining correct annotations and discarding incorrect ones. Finally, the morphs corresponding to each original document are the combination of the results from the previous manual annotation and this step of collaborative annotation.

\textbf{(1) Rule-based method}: Using the corpus manually annotated in the previous step, we constructed a morph dictionary $D$ whic contains 430 original words and their corresponding 2,688 morphs. Each entry in the dictionary contains one original word along with their multiple morph words, where the relationship between original word and morphs is one-to-many. 

During the annotated process automatically, we search each instance of the manually annotated corpus to find the morphs in the dictionary. If a match is found, this instance and the identified morph word will undergo further manual verification

\textbf{(2) Pre-trained language model based method}: Using the manually annotated corpus, we fine-tuned the pre-trained language model Mengzi-T5 \cite{zhang2021mengzi}. The details of the method is shown in section 5.1. During the annotated process automatically, each instance is input into the fine-tuned model, and the model's input and output were compared. If the input and output differed, it indicated that there might be omitted morph in the  sample. These samples were further examined, and upon confirmation, they were appropriately annotated. 

\begin{table}[!htbp]
\small
\centering
\begin{tabular}{l|c|c}
\hline
 & Positive\&Negative & Morph Num \\ \hline
Train & 6,236/76,554 & 7,301 \\ \hline
Valid & 800/800 & 1,025 \\ \hline
Test1 & 800/800 & 1,081 \\ \hline
Test2 & 400/400 & 548 \\ \hline
\end{tabular}
\caption{The statistics of the constructed Chinese morph dataset.}
\label{tab:2}
\end{table}

\textbf{Data Analysis:} Since the dataset construction is highly dependent on ASR outputs, the same speech input may produce different ASR results when processed by different ASR models. For example, the morph form \begin{CJK}{UTF8}{gbsn}‘‘白某障’’(\textit{b\'{a}i m\v{o}u zh\`{a}ng})\end{CJK} for \begin{CJK}{UTF8}{gbsn}‘‘白内障’’\end{CJK}(\textit{b\'{a}i n\`{e}i zh\`{a}ng, cataract}) could be transcribed as \begin{CJK}{UTF8}{gbsn}‘‘白母障’’(\textit{b\'{a}i m\v{u} zh\`{a}ng})\end{CJK}, \begin{CJK}{UTF8}{gbsn}‘‘白某张’’(\textit{b\'{a}i m\v{o}u zh\={a}ng})\end{CJK}, \begin{CJK}{UTF8}{gbsn}‘‘白某章’’(\textit{b\'{a}i m\v{o}u zh\={a}ng})\end{CJK} by different ASR models. 

To conduct a more comprehensive evaluation, We re-annotated the second test set (denoted Test2), selecting both different live streaming rooms and different ASR method. The Test2 includes 400 positive and negative instances.

Following the above process, we constructed a high-quality and comprehensive morph dataset, as shown in Table \ref{tab:2}. Dataset consists of 8,236 positive samples and 78,554 negative samples. The dataset includes a total of 431 original words and their corresponding 2,688 morphs forms, in which each word has nearly 7 morph words on average.

\section{Methods}

\textbf{LiveAMR method:} Existing morph resolution methods generally use non-autoregressive language model MacBERT, a corrective masked language model pre-training task was added to the BERT model \cite{wang2024cmright}. In the LiveAMR task, since the length of the variant words does not equal the length of the original word, we will use a text-to-text pre-trained model as a backbone, such as BART \cite{lewis2019bart} and Mengzi-T5 \cite{zhang2021mengzi}. Below are the steps involved in this process.

The created dataset consists of source-target pairs ($X$ and $Y$), where: \( X \) is the input text ( live stream transcript), \( Y \) is the desired output text (the normal text without morph words). The goal of the model is to learn a mapping from \( X \) to \( Y \).

The pre-trained model \( \mathcal{M} \) is a transformer-based sequence-to-sequence architecture, which is typically structured as: (1) Encoder: Takes the input sequence \( X \) and encodes it into hidden states; (2) Decoder: Takes the encoder's hidden states and generates the target sequence \( Y \).

During training, the model aims to minimize the loss, which is typically the Cross-Entropy Loss for text generation tasks. The formula for Cross-Entropy Loss is:

\[
\mathcal{L} = - \sum_{i=1}^{T} \sum_{v=1}^{V} \hat{y}_{i,v} \log p(y_{i,v} | X)
\]
where \( T \) is the length of the target sequence, \( V \) is the size of the vocabulary, \( \hat{y}_{i,v} \) is a one-hot encoding of the true token at position \( i \) in the target sequence, and \( p(y_{i,v} | X) \) is the predicted probability of token \( y_i \) at position \( i \) given the input \( X \).

During training, the model minimizes the loss function \( \mathcal{L} \) with respect to the model parameters \( \theta \) over multiple iterations (epochs):

\[
\theta^{\star} = \arg\min_{\theta} \mathbb{E}[ \mathcal{L}(X, Y; \theta) ]
\]
Where \( \mathbb{E} \) denotes the expectation over the training data, \( \mathcal{L}(X, Y; \theta) \) is the loss function dependent on the input \( X \), the target \( Y \), and the model parameters \( \theta \).

After fine-tuning, the model generates new outputs for unseen inputs. This is done by feeding the input \( X_{\text{input}} \) through the model to obtain the predicted sequence \( Y_{\text{pred}} \):

\[
Y_{\text{pred}} = \mathcal{M}(X_{\text{input}})
\]
Where \( Y_{\text{pred}} \) is the generated sequence, which can be decoded back into text.

\textbf{Data Augmentation via LLMs:} \label{llm} Some studies suggest that LLMs can be used to generate training datasets \cite{ding2023enhancing}. Although manual annotation can yield morph data from the real world, it comes at a high cost and may contain some redundancy, limiting the scale and diversity of the dataset. Therefore, we aim to leverage LLMs to generate more morph data to supplement manually annotated data and enhance the model's generalization ability. 

However, given the complexity of morph forms and the limitations of LLMs in understanding them, we did not directly ask the LLMs to generate sentences containing morphs. To this end, we propose a more reliable construction strategy that combines the annotated morphs lexicon with LLM capabilities. The specific steps are as follows:

(1) We randomly select a positive example from the training set and extract the corresponding morph words $WS$. There may be one or more morph words.

(2) Based on the morph dictionary $D$, we obtain the original word $WO$ for $WS$.

(3) We had the LLM simulate a live commerce scenario to generate 5 different sentences containing $WO$.

(4) According to the morph  dictionary $D$, we replace the original word $WO$ with different morph words to construct a set of sentences containing different morph words.

Through this approach, we constructed a manually created morph dataset containing 11,280 positive samples and 2,155 negative samples. Additionally, each positive sample generated by the LLM averages 2.87 morphs. This data effectively supplements the manually annotated data, increasing the scale and diversity of the model's training data. In Table \ref{example-morph}, show some specific examples.

\begin{table*}[!ht]
\small
\centering
\begin{tabularx}{\textwidth}{m{2cm}|XXXX|XXXX}
\hline
 & \multicolumn{4}{c|}{\textbf{Test1}} & \multicolumn{4}{c}{\textbf{Test2}} \\ \hline
\textbf{Method} & Acc & Pre & Recall & F1 & Acc & Pre & Recall & F1 \\ \hline
GPT & 0.405 & 0.421 & 0.320 & 0.364 & 0.496 & 0.494 & 0.441 & 0.466 \\
Deepseek & 0.605 & 0.660 & 0.529 & 0.587 & 0.677 & 0.667 & 0.626 & 0.646 \\
GLM & 0.451 & 0.484 & 0.515 & 0.499 & 0.532 & 0.525 & 0.649 & 0.580 \\
\hdashline[3pt/5pt]
Kenlm & 0.583 & 0.607 & 0.372 & 0.537 & 0.516 & 0.515 & 0.513 & 0.514 \\ 
Seq2Edit & 0.651 & 0.968 & 0.361 & 0.526 & 0.702 & 0.987 & 0.408 & 0.588 \\

Convseq2seq & 0.740 & 0.978 & 0.527 & 0.685 & 0.687 & 0.898 & 0.421 & 0.573 \\

BART  & 0.708 & 0.701 & 0.767& 0.738 & 0.656 & 0.670 & 0.611 & 0.639 \\
\hdashline[3pt/5pt]
T5 & 0.893 & \textbf{0.989} & 0.801 & 0.888 & 0.760 & \textbf{0.968} & 0.536 & 0.690 \\
+Aug & \textbf{0.928} & 0.937 & \textbf{0.927} & \textbf{0.932} & \textbf{0.863} & 0.929 & \textbf{0.787} & \textbf{0.852} \\ \hline
\end{tabularx}
\caption{The results of different methods,  where “+Aug” indicates fine-tuned the model using data augmentation via LLM.}
\label{experiemnt1}

\end{table*}

\section{Experiment}

\subsection{Experimental Setup}

\textbf{Metrics.} We expect the model to modify only the morphs in the target sentences without altering any other parts. A strict sentence-level assessment is applied: a positive sample is considered successfully predicted only when all morphs are correctly restored. For negative samples, a negative sample is deemed successfully predicted only if the model makes no modifications at all.

\textbf{Baselines.} The following models were  selected as the baseline for comparison:

(1)\textbf{LLMs}: To explore the morphs resolution capabilities of LLMs, we chose three representative models in the field of Chinese language understanding: GPT-3.5-turbo \footnote{https://openai.com/}, Deepseek -V2\footnote{https://platform.deepseek.com/}, and GLM4-Plus\footnote{https://chatglm.cn/}.  We manually selected 8 examples from the training set, including 6 positive samples and 2 negative samples, to be added as context to the prompt. The temperature was uniformly set to 0.7.

(2)\textbf{Seq2seq Model}: We selected two Seq2seq models Convseq2seq \cite{gehring2017convolutional} and BART \cite{lewis2019bart} as backbone, and fine-tune the model on the constructed training datset. 

(3)\textbf{Others}: To better illustrate that seq2seq is more suitable for the morph resolution task, we chose to analyze the statistical language model Kenlm \cite{heafield2011kenlm} and  BERT-based model Seq2Edit \cite{omelianchuk2020gector}.

(4) \textbf{Our method}: It is based on  T5 (mengzi-T5 \cite{zhang2021mengzi}). This model adopts the T5 training paradigm and has been retrained on large-scale Chinese corpora. \subsection{Implementation Details}
It is based on  T5 (mengzi-T5 \cite{zhang2021mengzi}). The Mengzi T5 model includes an encoder and decoder, where each consisting of 12 layers of Transformer layers. This model adopts the T5 training paradigm and has been retrained on large-scale Chinese corpora.

During the training process, the maximum length of the input sequence is set to 128, and the initial learning rate is set to 1e-4. We train the model for 20 epochs on a 24GB Nvidia 3090Ti GPU with the batch size set to 32. We use the AdamW optimizer, and the model employs a cosine annealing learning rate schedule.

\subsection{Experimental Results} 

The experimental results, presented in Table \ref{experiemnt1}, reveal that character-level correction methods like Seq2Edit and the statistical language model Kenlm are inadequate for addressing morphs in live streaming scenarios. In contrast, Seq2seq models (Convseq2seq, BART, and T5) perform better at managing inconsistencies in output length. Notably, the T5 model achieved the highest F1 score across both test sets, demonstrating its effectiveness for this task.

For T5 method, the results via data augmentation improved the F1 scores of T5 model by 4.95\% on Test1; on Test2, the improvements was 23.47\%. Our method shows stable performance across different test sets due to its contextual learning capabilities. On Test1, its performance is slightly lower than the baseline model, likely because the baseline excels with data similar to the training set. However, on Test2, which uses data from a different ASR model, the LLM's performance matches that of fine-tuned Seq2seq models, demonstrating its generalization ability with varied data distributions.

\subsection{Usefulness of Morph Resolution}

To investigate the role of morph resolution in detecting violations in e-commerce live streaming scenarios, we conducted a simple usability experiment.

\textbf{Setup.} We selected 4,641 live streaming clips for ASR processing and annotated the transcription results for each clip. After thorough consultation with market regulators, we have categorized the identification of violations in live-streaming sales videos into three types: compliance, suspected violation, and serious violation. Specifically, the "compliance" category refers to content that fully adheres to relevant regulations and platform rules, without any violation. The "suspected violation" category covers content that may potentially involve violation behaviors but requires further verification, such as suspected acts of inducing irrational consumption. The "serious violation" category pertains to actions that are explicitly prohibited by the platform or regulations, such as promoting healthcare products as drugs.

\begin{table}[]
\centering
\caption{Statistical information on dataset.}
\begin{tabular}{l|ll}
\hline
 & Class & Number \\ \hline
\multirow{3}{*}{Training set} & Compliance& 2,250 \\
& Suspected violation & 557 \\
& Serious Violation & 1,150 \\ \hline
\multirow{3}{*}{Validation Set} & Compliance& 130 \\
& Suspected violation & 130 \\
& Serious Violation & 130 \\ \hline
\multirow{3}{*}{Test set} & Compliance& 50 \\
& Suspected violation & 25 \\
& Serious Violation & 25 \\ \hline
\end{tabular}
\label{tab:statistic2}
\end{table}

We annotated a total of 4,447 instances including 2,430 compliances, 1,305 suspected violations, and 712 serious violations. We divided them into a training set, a validation set, and test set. The test set includes 100 samples, and the validation set contains 390 samples. The statistical information of the constructed CLiveSVD dataset is presented in Table \ref{tab:statistic2}.

\begin{table}[!h]
\begin{tabularx}{\columnwidth}{X|X|XXXX}
\hline
Method & Cat. & Acc & Pre & Recall & F1 \\ \hline
\multicolumn{1}{c|}{\multirow{3}{*}{Defalut}} & 0 & 0.81 & 0.917 & 0.88 & 0.89 \\
\multicolumn{1}{c|}{} & 1 & 0.81 & 0.77 & 0.68& 0.72 \\
\multicolumn{1}{c|}{} & 2 & 0.91 & 0.66 & 0.80 & 0.72 \\ \hline
\multirow{3}{*}{Morph} & 0 & 0.90 & 0.96 & 0.96 & 0.96 \\
 & 1 & 0.90 & 0.77 & 0.84& 0.80 \\
 & 2 & 0.90 & 0.91 & 0.84 & 0.87 \\ \hline
\end{tabularx}
\caption{Comparison of experimental results. "Default" indicates that the ASR results of the video are not processed. "Morph" refers to the processing of the ASR results for morph resolution. "0" represents compliant categories, "1" indicates suspected violation categories, and "2" denotes serious violation categories.}
\label{experiment2}
\end{table}

\textbf{Implements.}  It is important to note that in the default method, neither the training set nor the test set undergoes any changes, while in the comparison method, both the training set and the test set are processed with morph resolution. The BERT \cite{kenton2019bert} model was fine-tuned for classification task. 

\textbf{Results.} As shown in Table \ref{experiment2}, after resolution morphs in the original ASR results, the F1 scores for the compliant, suspected violation, and serious violation categories increased by approximately 6.91\%, 11.76\%, and 20.36\%, respectively, compared to the unprocessed results. This demonstrates that morph resolution can significantly improve the model's accuracy in detecting v.
\begin{figure}[!ht]
  \includegraphics[width=\columnwidth]{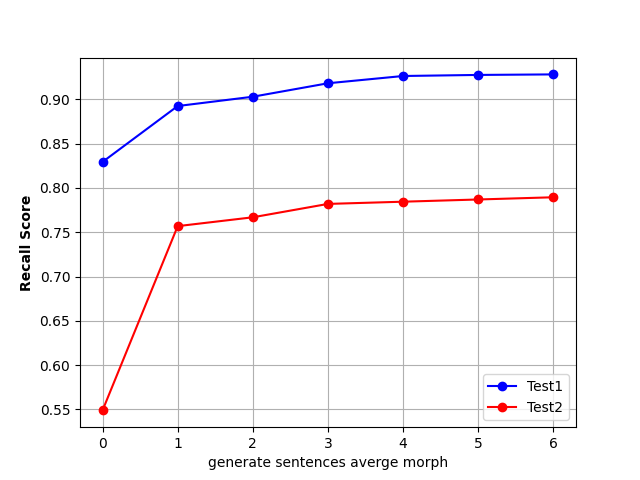}
  \caption{Performance with different number of training samples.}
  \label{ad1}
\end{figure}

\subsection{Ablation Study}

We explored the impact of data augmentation quantity on model performance. As shown in Section ~\ref{llm}, we controlled the data augmentation by setting the number of sentences generated for each original word. The sentence counts were set to 1, 2, 3, 4, 5, and 6, resulting in data volumes of 2,693, 5,373, 8,058, 10,744, 14,405, and 16,116, respectively.

In Figure \ref{ad1}, the experimental results show that data augmentation has a significant positive impact on model performance. At the same time, when the variable is set to 5, the number of augmented samples reaches 14,405, and the model's performance tends to stabilize.

\section{Conclusion}

This study introduces the task of morph resolution in live streaming scenarios, termed LiveAMR. A LiveAMR dataset was created through human-LLM collaboration, comprising 7,836 positive and 91,119 negative samples. The study analyzed task characteristics and utilized a text-to-text model architecture for morph resolution. Given the impracticality of manually constructing large-scale training corpora, an efficient data augmentation method based on LLMs was proposed, leveraging existing annotated data. Experimental results show that this augmentation method enhances model performance compared to baselines. The findings also indicate that morph resolution can contribute positively to streaming regulation.

\section*{Limitations}
We only annotated the live streaming domain where morphs are frequently used to evade censorship, without covering all topics in the live streaming field. Additionally, we validated the effectiveness of our proposed data augmentation method on only three models. In the future, we plan to expand this dataset and continue exploring the linguistic phenomenon of morphs.

\section*{Ethics Statement}
All data was collected from publicly available sources on the Douyin platform, ensuring no violation of privacy or data protection laws. Our aim is to address false advertising in health and medical live streams, contributing to consumer protection and industry standardization. Furthermore, this work serves the dual purposes of addressing moral concerns and navigating political censorship.

Human annotation was conducted by trained annotators who followed ethical guidelines, and we used large language models to enhance annotation accuracy. No personal or sensitive information was used, and all data was anonymized to prevent misuse.

Our findings support the development of tools to combat deceptive practices in e-commerce live streaming, ultimately benefiting consumers. The dataset and code will be made publicly available following ethical guidelines to encourage further research.

\section*{Acknowledgement}

This research is partially supported by  the National Language Commission of China (ZDI145-71), the National Natural Science Foundation of China (62076217), the Blue Project of Jiangsu and Yangzhou University, and the Top-level Talents Support Program of Yangzhou University.

\bibliography{custom}

\appendix

\section{The annotation Website}

\begin{figure}[!h]
  \includegraphics[width=\columnwidth]{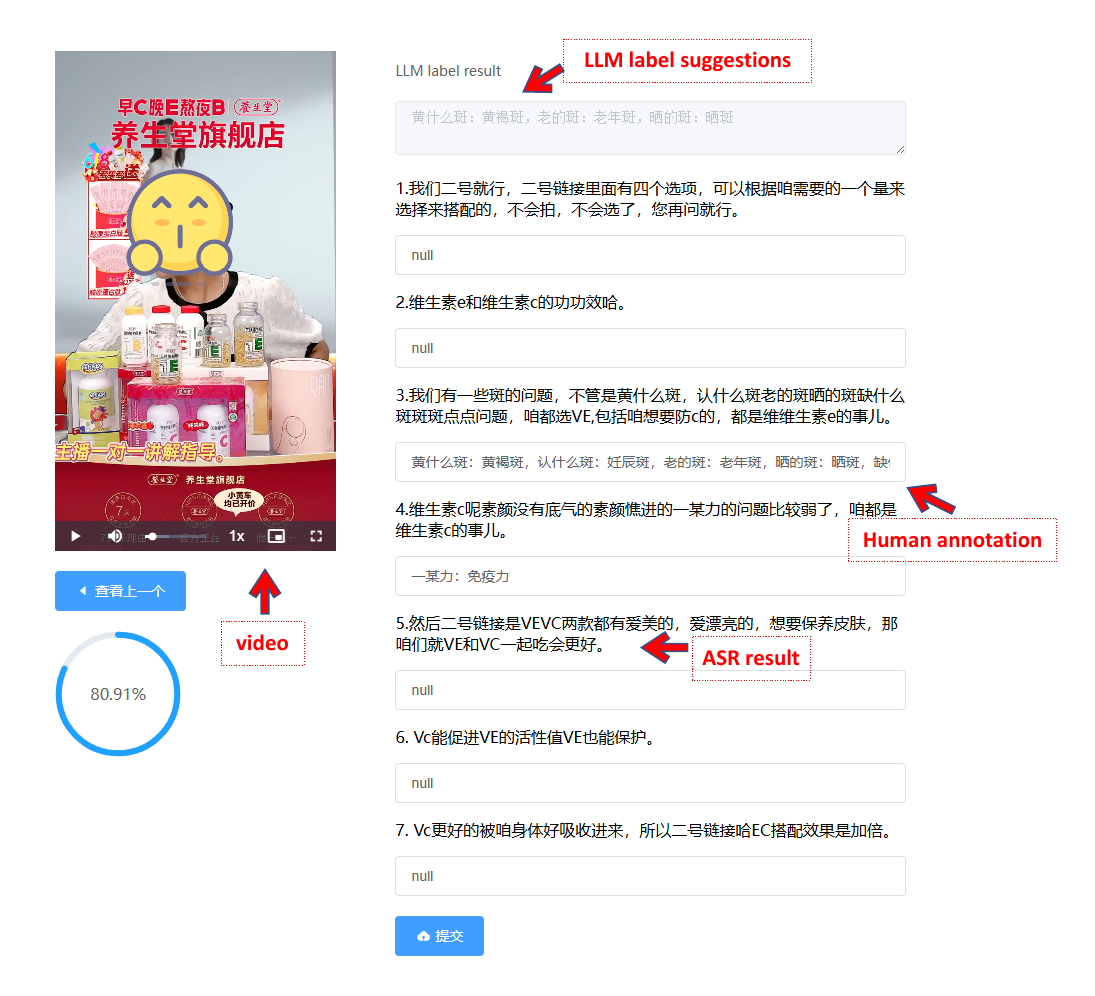}
  \caption{Screenshot of an annotation example on the annotation Website. The red text indicates added comments.}
  \label{fig:website}
\end{figure}

We have built a website based on Vue+FastAPI for annotators’ labeling work, as shown in Figure \ref{fig:website}. Due to the unique nature of the research scenarios, the annotators needed to process multiple modalities of information, which enhanced the quality and accuracy of the annotation results. At the same time, this is a time-consuming task, and we extend our sincerest gratitude to the annotators for their efforts.

\section{Prompt templates in this paper}

\textbf{ChatGPT-Generate Sentences.} The prompting template of ChatGPT-Generate sentences include targets words is shown in Figure~\ref{fig:ss}.

\begin{figure}[h!]
\centering
\begin{boxedminipage}{\columnwidth}
\footnotesize
Your role is that of a live-streaming host promoting products. You need to generate five promotional sentences that include the target words. Here are some real promotional sentences for you to mimic. The sentences should not have repeated meanings. The target word should remain unchanged. The length of the sentences should be as consistent as possible with the examples provided.\\
Target Words: \\
{[\textit{Target Words}]} \\
Examples: \\
{[\textit{Examples}]} \\
Generated Sentences:
\end{boxedminipage}
\caption{The prompting template of generating sentences. Generate context-appropriate sentences that contain the specified vocabulary and meet the required quantity.}
\label{fig:ss}
\end{figure}

\section{More Examples}
Here, we randomly some samples from morph dataset  in Table \ref{example-morph}.

\begin{table*}[!ht]
\begin{CJK}{UTF8}{gbsn}
\begin{tabularx}{\textwidth}{l|X}
\hline
 \textbf{Method}& \textbf{Sentence}  \\ \hline
\multicolumn{1}{l|}{\multirow{3}{*}{Real}} & \small{BC组合在三号选项三宝贝那维生c呢孩子，我们自己老年人\textcolor{red}{免某粒}特别弱，经常被其他人连带，经常\textcolor{red}{阿秋阿秋}的。} \newline
\small{The BC combination in option three significantly impacts children. Older adults have particularly weak immunity and often catch colds from others.}
\newline
\small{\textcolor{red}{免某粒(\textit{mi\v{a}n m\v{o}u l\`{i}:Free of certain pills})}:免疫力(\textit{mi\v{a}n y\`{i} l\`{i}, immunity})
\newline
\textcolor{red}{阿秋阿秋\textit{(\={a} qi\={u} \={a} qi\={u},Aqiu Aqiu})}:感冒(\textit{g\v{a}n m\`{a}o,catarrh})}
\\
\cdashline{2-2}
\multicolumn{1}{c|}{} & \small{都知道用\textcolor{red}{小蓝帽}什么意思吧，对不对？}
\newline
\small{You all know what the little blue hat means, right?}
\newline
\small{\textcolor{red}{小蓝帽(\textit{xi\v{a}o l\'{a}n m\`{a}o,small blue hat})}:保健食品标志(\textit{b\v{a}o ji\`{a}n sh\'{i} p\v{i}n bi\={a}o zh\`{i},Health Supplement Approval Mark})}
\\
\cdashline{2-2}
\multicolumn{1}{c|}{} & 
\small{我们一号链接三百一十八\textcolor{red}{米}，两桶。}
\newline
\small{Our link number one is 318 yuan, for two barrels.}
\newline
\small{\textcolor{red}{米(\textit{m\v{i},rice})}元(\textit{yu\'{a}n,yuan})}
\\ \hline

\multirow{3}{*}{LLM} & 
\small{想要\textcolor{red}{改某善}身体\textcolor{red}{某平某衡}？试试我们的新品，今天下单有特别优惠，立减50\textcolor{red}{米}！}
\newline
\small{Want to improve your balance? Try our new product, order today for a special discount of 50 yuan off!}
\newline
\small{\textcolor{red}{改某善(\textit{g\v{a}i m\v{o}u sh\`{a}n,improvement})}:改善(\textit{g\v{a}i  sh\`{a}n,improvement})
\newline
\textcolor{red}{某平某衡(\textit{m\v{o}u p\'{i}ng m\v{o}u h\'{e}ng,balance})}:平衡(\textit{p\'{i}ng h\'{e}ng,balance})
\newline
\textcolor{red}{米(\textit{m\v{i},rice})}元(\textit{yu\'{a}n,yuan})}
\\
\cdashline{2-2}
 &  
 \small{我们的产品专为\textcolor{red}{孕妈妈}设计，能够帮助控制\textcolor{red}{糖高}，减轻身体\textcolor{red}{猛副某用}，让孕期更加轻松。}
 \newline
 \small{Our products are designed specifically for pregnant women to help control hyperglycemia and relieve certain body effects, making pregnancy easier.}
 \newline
 \small{\textcolor{red}{孕妈妈(\textit{y\`{u}n m\={a} m\={a},Pregnant mother})}:孕妇(\textit{y\`{u}n f\`{u},pregnant})
  \newline
 \textcolor{red}{糖高(\textit{t\'{a}ng g\={a}o,high in sugar})}:高血糖(\textit{g\={a}o xu\`{e} t\'{a}ng,hyperglycemia})
\newline\textcolor{red}{猛副某用(\textit{m\v{e}ng f\`{u} m\v{o}u y\`{o}ng,side effect})}:副作用(\textit{f\`{u} zu\`{o} y\`{o}ng,side effect})}
 \\
 \cdashline{2-2}
 & 
 \small{\textcolor{red}{运和动}不仅有助于心血管健康，还能减少\textcolor{red}{某血某栓}形成的风险，\textcolor{red}{百大褂}也经常强调这一点。}
 \newline
 \small{Exercise not only helps cardiovascular health, but also reduces the risk of thrombus, which doctors often emphasize.}
 \newline
 \small{\textcolor{red}{运和动(\textit{y\`{u}n h\'{e} d\`{o}ng,movement and motion})}:运动(\textit{y\`{u}n d\`{o}ng,exercise})
 \newline
 \textcolor{red}{某血某栓(\textit{m\v{o}u xu\`{e} m\v{o}u shu\={a}n,thrombus})}：血栓(\textit{xu\`{e} shu\={a}n,thrombus})
 \newline
 \textcolor{red}{百大褂(\textit{b\v{a}i d\`{a} gu\`{a},people in white})}:医生(\textit{y\={i} sh\={e}ng,doctor})}
 \\ \hline
\end{tabularx}
\end{CJK}
\caption{Morph sample display: The first row contains sentences with morphs, the second row is the translation, and the third row shows the morph annotation results. "Real" indicates that the data source is real data, not synthetic data. "LLM" indicates data synthesized using an LLM-based method, shown in ~\ref{llm}.}
\label{example-morph}
\end{table*}

\end{document}